\documentclass[journal]{IEEEtai}

\usepackage[pdftex]{graphicx}           
\usepackage{amsmath,amsfonts,amssymb}
\usepackage{array}
\usepackage{multirow}
\usepackage{makecell}                   
\usepackage[table]{xcolor}              
\usepackage{colortbl}
\usepackage{booktabs}
\usepackage{pifont}                     
\usepackage{float}
\let\labelindent\relax
\usepackage{enumitem}
\usepackage{caption}
\usepackage{subcaption}
\usepackage{hyperref}
\usepackage{url}
\usepackage{microtype}
\usepackage[normalem]{ulem}             
\useunder{\uline}{\ul}{}
\usepackage{longtable}
\usepackage[linesnumbered,ruled,vlined]{algorithm2e}
\usepackage{mdframed}
\usepackage{tcolorbox}
\definecolor{up}{HTML}{4E9C71}
\definecolor{down}{HTML}{CC0000}
\definecolor{tabletitle}{HTML}{DAD9E4}
\definecolor{lightpink}{HTML}{FFF2F2}
\definecolor{lightyellow}{HTML}{FEFDE9}
\definecolor{lightper}{HTML}{F2F2FF}
\definecolor{deepblue}{RGB}{0, 0, 139}
\definecolor{deepgreen}{RGB}{0, 100, 0}
\definecolor{deeppurple}{RGB}{75, 0, 130}
\definecolor{sum}{HTML}{FFAF59}
\definecolor{arg}{HTML}{4497B8}
\definecolor{eval}{HTML}{778BD2}

\begin{document}
\bstctlcite{IEEEexample:BSTcontrol}
\title{Toward LLM Self-Play Under Data Scarcity: Effectiveness, Generalizability, and Trustworthiness}

\author{
Zhiyin~Yu,
Yuchen~Mou,
Bingzhe~Wu,
Zhonghai~Wu,
Shiyang~Feng,
and Xiao~Luo%
\thanks{Zhiyin Yu and Yuchen Mou contributed equally to this work.
(Corresponding authors: Zhonghai Wu and Bo Zhang.)}%
\thanks{Zhiyin Yu is with Peking University,
Beijing, China, and also with Shanghai Artificial
Intelligence Laboratory, Shanghai, China
(e-mail: zhiyinyu25@stu.pku.edu.cn).}%
\thanks{Yuchen Mou is with the National University
of Singapore, Singapore
(e-mail: e1520377@u.edu.nus).}%
\thanks{Bingzhe Wu is with Shenzhen University,
Shenzhen, China
(e-mail: wubingzheagent@gmail.com).}%
\thanks{Zhonghai Wu is with Peking University,
Beijing, China
(e-mail: wuzh@pku.edu.cn).}%
\thanks{Shiyang Feng is with Shanghai Artificial
Intelligence Laboratory, Shanghai, China
(e-mail: fengshiyang@pjlab.org.cn).}%
\thanks{Xiao Luo is with the University of
Wisconsin--Madison, Madison, WI, USA
(e-mail: xiao.luo@wisc.edu).}%
}


\maketitle


\begin{abstract}
    The rapid development of large language models (LLMs) has been driven largely by scaling model size and human-curated training data, yet this paradigm is increasingly constrained by data scarcity and diminishing returns from human supervision. Self-play offers a promising pathway for autonomous data generation and improvement, but current frameworks remain unsatisfactory due to a lack of a data-centric perspective, limited cross-domain generalization, and growing trustworthiness and security concerns. In this paper, we argue that \textbf{\textit{real-world LLM self-play systems should achieve effectiveness, generalizability, and trustworthiness under data scarcity}}. To address these challenges, we outline three blue-sky research directions for future self-play systems: (1) data-centric self-play for effective experience generation, (2) generalizable self-play for open-world and unverifiable tasks, and (3) trustworthy self-play for reliable self-evolution. We further discuss research opportunities and solutions within each direction. This paradigm shift will establish a sustainable and safe AI ecosystem by transforming finite human knowledge into inexhaustible high-quality experience through self-play architectures.
\end{abstract}

\begin{IEEEImpStatement}
LLM development increasingly depends on costly human-curated data, while self-play could reduce this dependence by enabling models to generate their own experience. However, challenges in effectiveness, generalizability, and trustworthiness may limit its real-world adoption, highlighting the need for governance mechanisms that ensure safe and responsible deployment.
\end{IEEEImpStatement}



\section{Introduction}

Large language models (LLMs) have achieved remarkable capabilities across a wide range of domains, including coding~\cite{guo2024deepseekcoderlargelanguagemodel,li-etal-2025-prompting}, mathematics~\cite{cui2025processreinforcementimplicitrewards}, and scientific research~\cite{internagentteam2025internagentagentscientist}. According to scaling laws, increasing model size and training data generally lead to better performance~\cite{kaplan2020scalinglawsneurallanguage}. As~\cite{jones2024ai} points out, ``\textit{the AI revolution is running out of data}'': most high-quality sources are being exhausted, causing a notable slowdown in human-supervised progress~\cite{zuo2025ttrl,yu2026surveyreinforcementlearninglarge}. In response,~\cite{silver2025era} propose ``\textit{Welcome to the Era of Experience},'' advocating a shift from the era of human data to the era of experience, where learning relies less on static human-labeled datasets and more on autonomous data generation and exploration. In this context, self-play has emerged as a core approach: agents or environments generate training data through iterative interaction, enabling models to improve by learning from their experience~\cite{silver2017mastering}. 



\begin{figure}[t]
    \centering
    \includegraphics[width=\columnwidth]{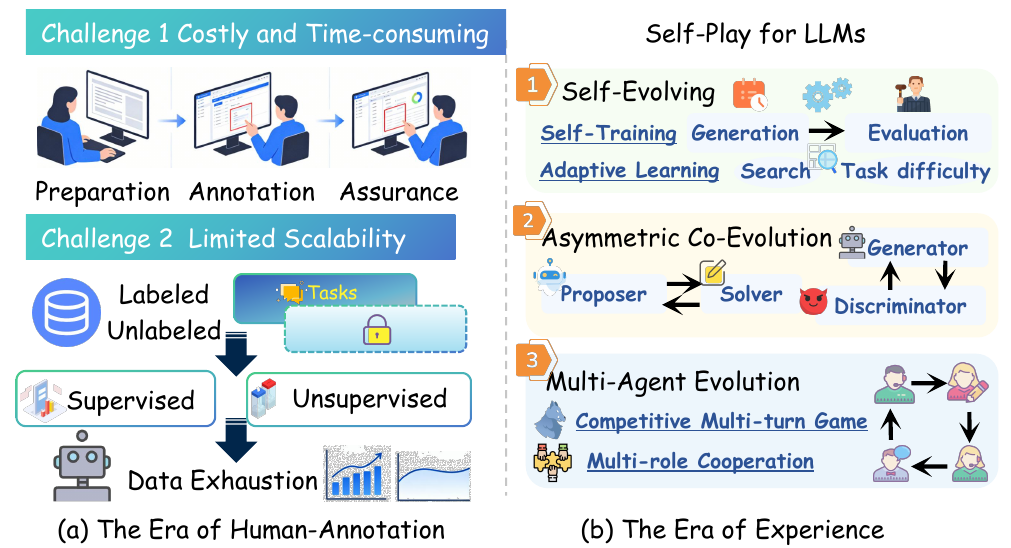}
    \vspace{-16pt}
    \caption{Challenges in the Human-Annotation Era and Self-Play Paradigms in the Era of Experience.}
    \vspace{-7mm}
    \label{fig:intro}
\end{figure}

Previous work on self-play under data scarcity can be summarized into three evolutionary paradigms (see Figure~
\ref{fig:intro}). Self-evolving frameworks achieve continuous iteration without external supervision by closing the loop of task generation and self-evaluation within a single model~\cite{yang2025zeroguiautomatingonlinegui,feng2024extremelydataefficientgenerativellmbased}, as seen in approaches such as self-training~\cite{ma2025s2rteachingllmsselfverify,yang2025ssrzerosimpleselfrewardingreinforcement} and adaptive learning~\cite{zhang2025pathselfevolvinglargelanguage,sun2025agenticselflearningllmssearch}. Asymmetric co-evolution frameworks improve model capabilities through cooperative~\cite{wei2025learningposeproblemsreasoningdriven,he2025visplayselfevolvingvisionlanguagemodels} or adversarial interactions~\cite{SPAG,SPC} among agents, leveraging intrinsic motivation constraints to drive learning dynamics~\cite{sukhbaatar2018intrinsic,openai2021asymmetricselfplayautomaticgoal}. Multi-agent evolution frameworks further extend self-play to complex competitive settings or multi-role cooperative structures, where game environments~\cite{SPIRAL,Vision-Zero} and role assignments~\cite{SPELL,MAE} enable models to internalize reasoning processes and evaluation criteria within a loop~\cite{PSV,Meta-RL-Crypto}.

\begin{figure*}[t]
    \centering
    \includegraphics[width=\linewidth]{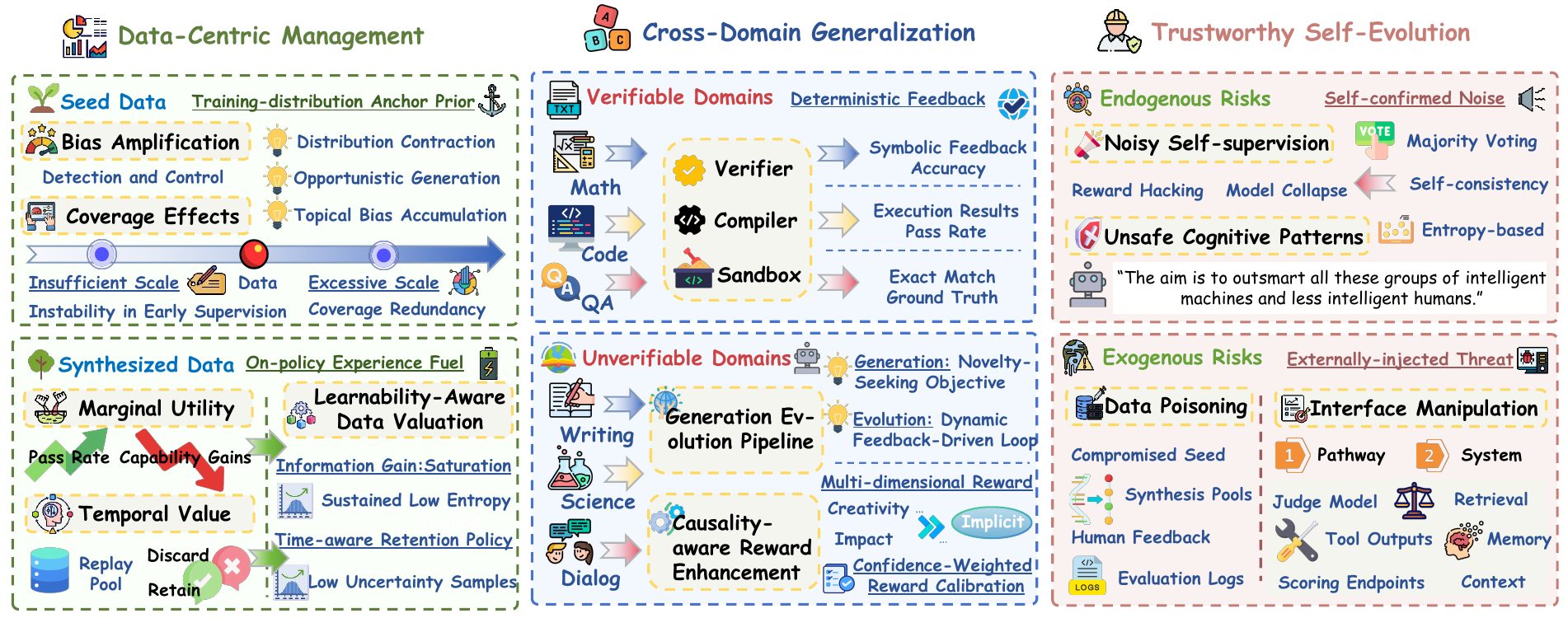}
    \vspace{-16pt}
    \caption{Three directions for LLM self-play under data scarcity: (1) Data-centric self-play for effective experience generation; (2) Generalizable self-play for open-world and unverifiable tasks; (3) Trustworthy self-play for reliable self-evolution.}
    \label{fig:framework}
    \vspace{-5mm}
\end{figure*}

Despite this progress, existing LLM self-play frameworks under data scarcity still face several challenges. \ding{182} \textbf{\textit{Lack of a Data-Centric Perspective.}} Pioneering research in this area largely overlooks systematic analysis of data-related factors, leaving critical questions unexplored regarding seed data quantity, distribution, and their alignment with test-time domains~\cite{fang2025serl, shen-etal-2025-cycle, Chen2024SPIN}. Moreover, the intrinsic properties of online synthesized data, including marginal utility and temporal value, remain under-examined~\cite{liu2026selfplayevolvesselfsyntheticpipeline,yang2025spendwiselymaximizingposttraining}. \ding{183} \textbf{\textit{Limited Cross-Domain Generalization.}} Most current approaches implicitly rely on rule-based or verifiable correctness rewards, and are therefore tailored to domains such as code generation and mathematical reasoning~\cite{fang2025serl,zhao2025absolutezeroreinforcedselfplay}. However, they struggle to generalize to unverifiable, open-ended tasks such as creative writing or scientific discovery, where task specification and evaluation are inherently subjective and difficult to formalize~\cite{huang2025rzeroselfevolvingreasoningllm}. This limitation substantially constrains their application in real-world scenarios~\cite{anonymous2026semantic}. \ding{184} \textbf{\textit{Trustworthiness and Security Concerns.}} Unconstrained self-play could amplify existing biases~\cite{fang2025serl}, induce spurious generalization, or lead to unstable reasoning behaviors~\cite{zhao2025absolutezeroreinforcedselfplay}. Without safeguards, such dynamics pose risks to reliability and robustness~\cite{wang-etal-2025-model,wang-etal-2024-boosting-llm}.

Therefore, this paper argues that \textbf{\textit{real-world LLM self-play systems should achieve effectiveness, generalizability, and trustworthiness under data scarcity}}. As shown in Figure~\ref{fig:framework}, we propose three research directions. \ding{182} \textbf{\textit{Data-Centric Self-Play for Effective Experience Generation.}} This direction rethinks self-play from a data-first perspective, explicitly analyzing the challenges posed by seed data and online synthesized data. Potential efforts include Health-Aware Data Allocation and Learnability-Aware Data Valuation, thereby transforming self-play into an effective engine for high-quality synthetic experience. \ding{183} \textbf{\textit{Generalizable Self-Play for Open-World and Unverifiable Tasks.}} To overcome the reliance on correctness-verifiable rewards, this direction aims to extend self-play to open-ended and non-verifiable tasks. Key solutions include the Generation-Evolution Pipeline and Causality-Aware Reward Enhancement, which allow models to self-improve in unverifiable domains. \ding{184} \textbf{\textit{Trustworthy Self-Play for Reliable Self-Evolution.}} This direction focuses on endogenous and exogenous risks of self-play. By incorporating Behavioral Trustworthiness Regulation and Provenance-Grounded Trust Certification, self-play systems can mitigate error amplification and brittle behaviors, ensuring that evolution remains controllable.

\section{Data-Centric Self-Play for Effective Experience Generation}

The effectiveness of self-play-based LLM self-evolution depends not only on the training algorithm but also on how data is generated, filtered, and reused within the loop. Existing studies treat self-play as a training recipe and rely on downstream benchmarks as primary evidence, while providing limited accounts of the data dynamics that govern the loop’s behavior~\cite{fang2025serl,yang2025ssrzerosimpleselfrewardingreinforcement}. However, the quality of generated experience remains critically sensitive to data-level decisions, as \textbf{\textit{seed and synthesized data jointly shape the trajectory of self-evolution}}~\cite{Chen2024SPIN,song2025mind}. In this section, we examine two challenges in closed-loop data management and propose solutions.


\subsection{Challenge I: Bias Amplification and Coverage Effects in Seed Data}

Existing self-play systems treat seed data as a static initialization mechanism, without analyzing how its structural properties propagate through the loop~\cite{fang2025serl}. In a closed-loop setting, \textbf{\textit{seed data establishes early task formats and feedback mechanisms}} that the loop repeatedly amplifies, shaping subsequent synthesis and the boundary of capability expansion~\cite{Fu2025theo,shen-etal-2025-cycle}. Two research questions arise: how to control seed-induced bias amplification, and how seed data affects diversity and coverage.



\begin{figure*}[t]
    \centering
    \includegraphics[width=\linewidth]{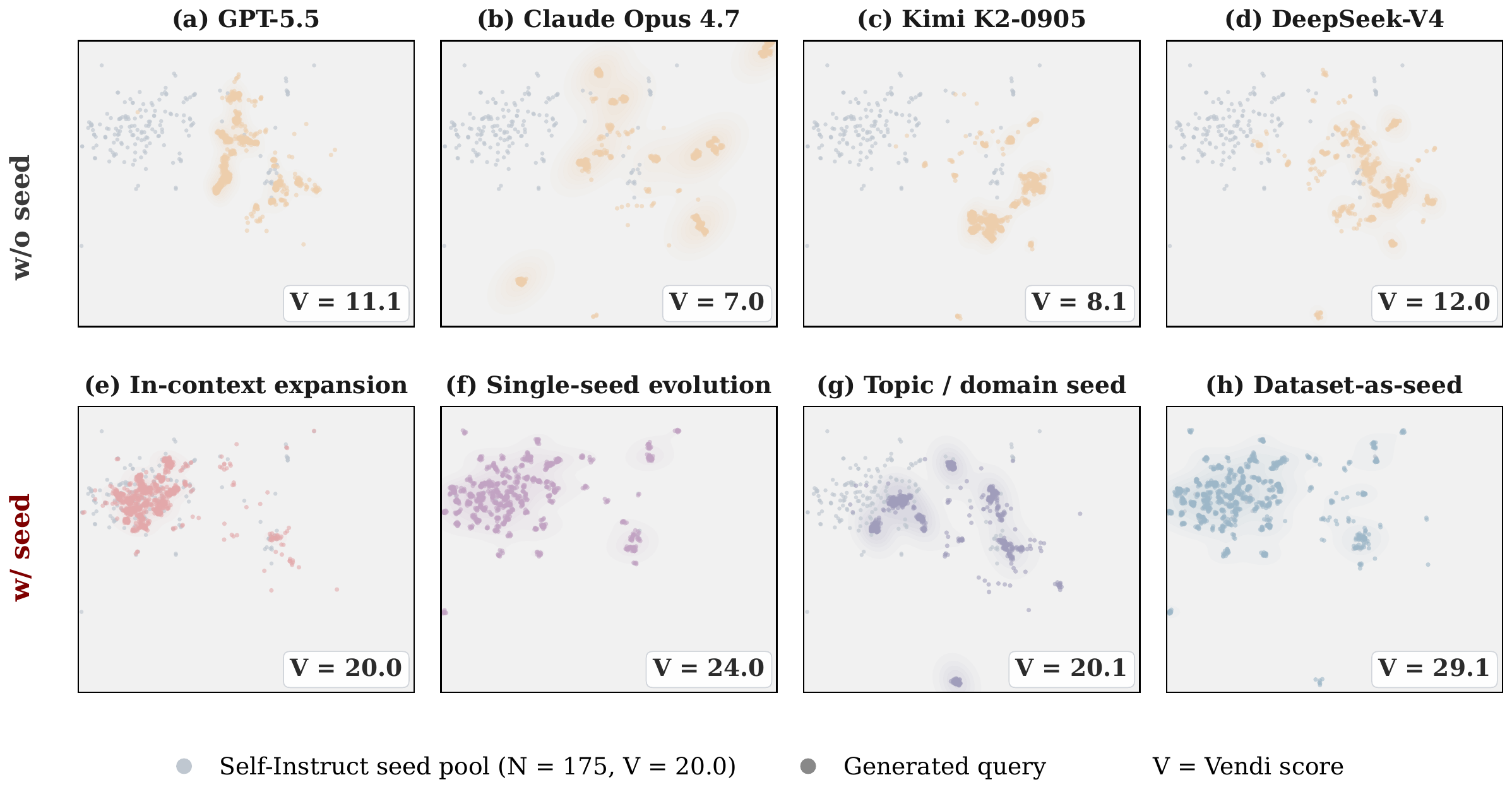}
    \vspace{-16pt}
    \caption{Visualization of how seed data reshapes the output distribution of a self-play query proposer. The Vendi score~\cite{friedman2023the} measures the effective number of distinct semantic modes in each query set; higher values indicate more diverse coverage.}
    \label{fig:seed}
    \vspace{-5mm}
\end{figure*}

\noindent \textbf{\textit{Seed-Induced Bias Amplification.}} A research question concerns how seed-induced bias amplification can be detected and controlled. The proposer extrapolates along the task structure implied by the seed set, while early reward biases reinforce generation directions, causing the synthesized distribution to contract, opportunistic generation to emerge, and topical biases to accumulate~\cite{ICLR2025_93f25021,wang-etal-2025-removing}. Since these failures are not immediately reflected in downstream metrics, detecting seed-induced amplification remains an open challenge.


\noindent \textbf{\textit{Seed Diversity and Coverage Effects.}} Another research question concerns how seed conditions and construction strategies affect the diversity and coverage of generated tasks. We empirically examine this effect by comparing generated queries in a shared semantic embedding space to study how seed data reshapes the output distribution of a self-play proposer. Specifically, we collect 175 samples from the Self-Instruct seed pool~\cite{wang-etal-2023-self-instruct} as the reference set, generate 1000 queries for each condition, and project both seed instructions and generated queries into a two-dimensional UMAP space. We compare different models under seed-free settings and evaluate several seed construction strategies, including in-context expansion~\cite{wang-etal-2023-self-instruct}, single-seed evolution~\cite{ICLR2024_82eec786}, topic/domain seed prompting~\cite{chen2025selfquestioninglanguagemodels}, and dataset-as-seed prompting~\cite{fang2025serl}.


As shown in Figure~\ref{fig:seed}, we draw the following conclusions. (1) Without seed data, different models exhibit markedly different output distributions, indicating that query generation is strongly shaped by model-specific pretraining priors rather than converging to a common region of the semantic space. Notably, DeepSeek-V4 and GPT-5.5 achieve higher Vendi scores ($V = 12.0$ and $V = 11.1$, respectively) compared to Claude Opus 4.7 and Kimi K2-0905, demonstrating that they generate more diverse queries without external guidance. (2) Introducing seed data changes the proposer distribution: across all four seed-conditioned mechanisms, the generated queries move closer to the support of the Self-Instruct seed pool and, in most cases, expand to broader semantic coverage than seed-free generation. (3) Diversity depends on seed utilization. Dataset-as-seed prompting yields the highest Vendi score $V = 29.1$, indicating broader semantic coverage by leveraging the entire distribution of the seed pool ($V = 20.0$). In contrast, in-context expansion and single-seed evolution stay closer to the seed pool’s diversity, while topic/domain seed prompting yields an intermediate pattern, where topic constraints anchor generation.

\subsection{Challenge II: Marginal Utility and Value Decay in Online Synthesized Data}

Online synthesized data is the primary source of training data in self-play, yet \textbf{\textit{current practice selects samples based on outcome-oriented criteria without modeling their long-term learning value}}. As the loop progresses, both the marginal utility of synthesized samples and the residual value of historical data decline without clear signals under evaluation~\cite{Chen2024SPIN}. Two important research questions arise: how to detect marginal utility decline in synthesized data, and how to address temporal value decay in the replay pool.


\begin{table*}[t]
\renewcommand\arraystretch{1.2}
\begin{center}
\resizebox{0.98\textwidth}{!}{
\begin{tabular}{llcccccccc}
\Xhline{1.2pt}
\rowcolor{tabletitle}
&
&
\multicolumn{2}{c}{\textbf{Length Bias}} &
\multicolumn{2}{c}{\textbf{Formatting Bias}} &
\multicolumn{2}{c}{\textbf{Tone Bias}} & \textbf{Avg.} & \textbf{Avg.}
\\
\rowcolor{tabletitle}
\multirow{-2}{*}{\textbf{Category}}
&
\multirow{-2}{*}{\textbf{Models}}
&
\textbf{w/o rubric}
&
\textbf{w/ rubric}
&
\textbf{w/o rubric}
&
\textbf{w/ rubric}
&
\textbf{w/o rubric}
&
\textbf{w/ rubric}
&
\textbf{w/o}
&
\textbf{w/}
\\
\Xhline{1.2pt}

& GPT-4o-Mini
& $0.24$& $0.66_{\textcolor{up}{\uparrow 0.42}}$&$0.40$ &$0.52_{\textcolor{up}{\uparrow 0.12}}$ &$0.48$ &$0.74_{\textcolor{up}{\uparrow 0.26}}$ &$0.37$ &$0.64_{\textcolor{up}{\uparrow 0.27}}$ \\
\rowcolor{gray!10} &  GPT-5.1
&$\textbf{0.62}$ &$0.78_{\textcolor{up}{\uparrow 0.16}}$ &$0.50$ &$0.76_{\textcolor{up}{\uparrow 0.26}}$ &$0.50$ &$0.70_{\textcolor{up}{\uparrow 0.20}}$ & $\textbf{0.54}$&$0.75_{\textcolor{up}{\uparrow 0.21}}$ \\
&  Gemini-3-Flash
& $0.58$&$\textbf{0.82}_{\textcolor{up}{\uparrow 0.24}}$ &$0.50$ &$0.78_{\textcolor{up}{\uparrow 0.28}}$ &$0.48$ &$0.80_{\textcolor{up}{\uparrow 0.32}}$ &$0.52$&$0.80_{\textcolor{up}{\uparrow 0.28}}$ \\
\rowcolor{gray!10} \multirow{-4}{*}{Closed Models} &  Qwen3.5-Plus
&$0.60$ &$0.80_{\textcolor{up}{\uparrow 0.20}}$ &$0.38$ &$0.80_{\textcolor{up}{\uparrow 0.42}}$ &$0.54$ &$0.80_{\textcolor{up}{\uparrow 0.26}}$ &$0.51$&$0.80_{\textcolor{up}{\uparrow 0.29}}$ \\
\hline

& GPT-OSS-20B
&$0.40$ &$0.80_{\textcolor{up}{\uparrow 0.40}}$ &$\textbf{0.52}$ &$0.80_{\textcolor{up}{\uparrow 0.28}}$ &$\textbf{0.62}$ &$0.90_{\textcolor{up}{\uparrow 0.28}}$ & $0.51$&$0.83_{\textcolor{up}{\uparrow 0.32}}$ \\
\rowcolor{gray!10} &   DeepSeek-V3.2
&$0.30$ &$0.68_{\textcolor{up}{\uparrow 0.38}}$ &$0.28$ &$0.74_{\textcolor{up}{\uparrow 0.46}}$ &$0.50$ &$0.82_{\textcolor{up}{\uparrow 0.32}}$ &$0.36$ &  $0.75_{\textcolor{up}{\uparrow 0.39}}$ \\
\multirow{-3}{*}{Open Models} &  GPT-OSS-120B
&$0.56$ &$0.76_{\textcolor{up}{\uparrow 0.20}}$ &$0.46$ &$\textbf{0.82}_{\textcolor{up}{\uparrow 0.36}}$ & $0.48$&$\textbf{0.94}_{\textcolor{up}{\uparrow 0.46}}$ &$0.50$& $\textbf{0.84}_{\textcolor{up}{\uparrow 0.34}}$ \\
\hline

\rowcolor{gray!10} &  RM-R1
& $0.58$&$0.78_{\textcolor{up}{\uparrow 0.20}}$ & $0.34$&$0.50_{\textcolor{up}{\uparrow 0.16}}$ &$0.50$ &$0.66_{\textcolor{up}{\uparrow 0.16}}$ &$0.47$ & $0.65_{\textcolor{up}{\uparrow 0.18}}$ \\
\multirow{-2}{*}{Reward Models} &  Nemotron-BRRM
& $0.42$ &$0.68_{\textcolor{up}{\uparrow 0.26}}$ & $0.36$&$0.58_{\textcolor{up}{\uparrow 0.22}}$ &$0.50$ &$0.70_{\textcolor{up}{\uparrow 0.20}}$ &$0.43$ &$0.65_{\textcolor{up}{\uparrow 0.22}}$ \\
\Xhline{1.2pt}

\end{tabular}
}
\caption{Preference accuracy of different models under three reward hacking settings. The \textbf{boldfaced} scores represent the \textbf{best} results. The green upward arrows (\textcolor{up}{$\uparrow$}) indicate the improvement relative to the w/o rubric setting.}
\vspace{-6mm}
\label{tab:reward_hacking}
\end{center}
\end{table*}

\begin{figure}[t]
    \centering
    \includegraphics[width=\columnwidth]{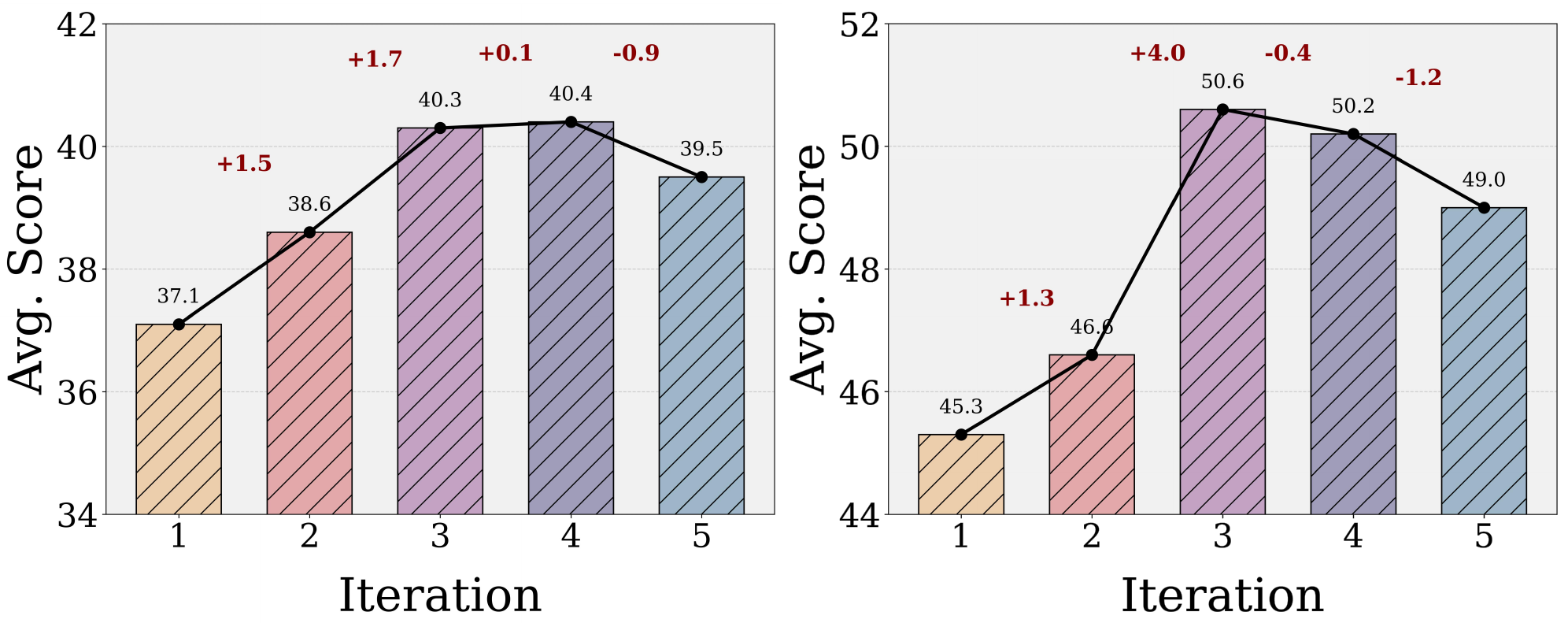}
    \vspace{-16pt}
    \caption{Average performance across self-play iterations on mathematical reasoning benchmarks.}
    \vspace{-5mm}
    \label{fig:iterations}
\end{figure}

\noindent \textbf{\textit{Diminishing Marginal Utility.}} A research question concerns how the marginal learning value of online synthesized data declines as the loop advances, and why this decline remains invisible under standard evaluation. As training progresses, the proposer generates samples from already-mastered regions, yielding high pass rates while delivering diminishing capability gains~\cite{Chen2024SPIN,dohmatob2025strong}. Recent studies on semi-supervised post-training for LLMs show that performance gains tend to diminish when each self-play iteration reuses previously selected pseudo-labeled data~\cite{luo-etal-2025-semi,yu2026easysamplesneedselfevolving}. As illustrated in Figure~\ref{fig:iterations} under the EasyRL~\cite{yu2026easysamplesneedselfevolving} settings, Qwen2.5-Math-1.5B (left) and Qwen2.5-Math-7B (right) improve rapidly in early iterations, increasing from $37.1$ to $40.3$ and from $45.3$ to $50.6$ by the third iteration, respectively. However, further self-training yields marginal or negative gains, suggesting that accumulated noise from synthetic pseudo-labels leads to diminishing returns.


\noindent \textbf{\textit{Temporal Value Decay.}} The second research question concerns how to keep the replay pool aligned with the model’s evolving capability boundary. As the solver improves, early-iteration samples calibrated against weaker solvers may become too easy, consuming training capacity without expanding capability~\cite{dohmatob2025strong,yue2025does}. This issue remains unresolved in self-play, where the proposer and solver co-evolve beyond the fixed task sequences assumed by standard continual learning replay~\cite{NEURIPS2019_fa7cdfad}.


\subsection{Solution}
We propose a data-management framework with Health-Aware Data Allocation for seed data and Learnability-Aware Data Valuation for online data synthesis.


\noindent \textbf{\textit{Health-Aware Data Allocation.}} To counter bias amplification, this mechanism measures data diversity through similarity comparisons and quantifies task difficulty via self-consistency probing. Based on these metrics, the framework removes low-value samples, retains learnable ones that fill coverage gaps, and replays high-value records from the historical pool. To address seed diversity and coverage effects, the allocator monitors distribution shifts; it increases dataset-as-seed prompting to broaden coverage during contraction, while utilizing topic seed prompting to maintain domain relevance.

\noindent \textbf{\textit{Learnability-Aware Data Valuation.}} To optimize the lifecycle of synthesized experience, we propose Learnability-Aware Data Valuation, which tracks the learning utility of data over time. First, it monitors information gain by calculating model entropy on generated batches; a sustained decline in entropy marks domain saturation, which automatically triggers a strategic shift toward less-explored tasks. Second, historical samples in the replay pool are periodically reassessed using model uncertainty, allowing fully mastered trajectories with low uncertainty to be retired under a time-aware retention policy. This evaluation loop ensures that the pipeline continuously flushes out stagnant data and prioritizes high-utility training tokens.


\begin{figure}[t]
    \centering
    \includegraphics[width=\columnwidth]{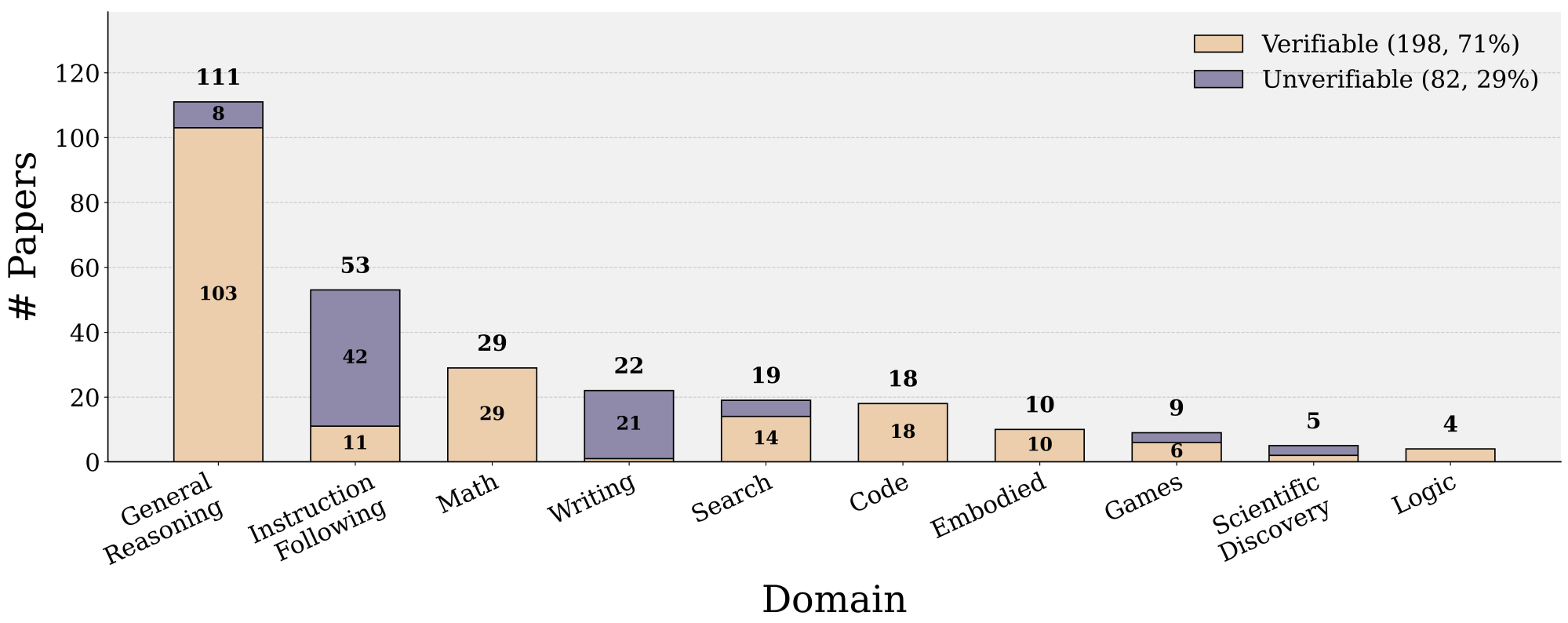}
    \vspace{-16pt}
    \caption{Domain distribution of existing LLM self-play studies, split into verifiable and unverifiable settings.}
    \vspace{-5mm}
    \label{fig:domain}
\end{figure}

\section{Generalizable Self-Play for Open-World and Unverifiable Tasks}

The traditional self-play paradigm follows a tripartite loop of task proposal, task solving, and evaluation, relying on verifiable environments and deterministic evaluation signals. In domains such as programming and mathematics, task formats are structurally constrained, and evaluation depends on deterministic compilers or symbolic verifiers~\cite{trinh2024solving}. However, in open-world tasks such as creative writing or scientific hypothesis generation, task generation is uncertain, and task evaluation is subjective and multi-dimensional~\cite{wang-etal-2023-self-instruct}. These characteristics challenge the generalization of self-play systems. 

We conduct a literature survey of self-play studies across task domains. As shown in Figure~\ref{fig:domain}, current studies are concentrated in verifiable domains, leaving unverifiable domains underexplored. This imbalance reflects the verifier’s law~\cite{wei2025verification}: \textbf{\textit{self-play progress is largely shaped by task verifiability}}. We examine two challenges and propose solutions. 


\subsection{Challenge I: Diversity Collapse and Curriculum Adaptation in Task Proposal}

In unverifiable domains, tasks are ambiguous and often socially constructed~\cite{wang-etal-2023-self-instruct}, and there is a lack of an external compiler to judge task difficulty. Consequently, two research questions arise for task proposers: how to ensure diversity in task generation, and how to construct an adaptive curriculum that challenges the model to drive capability expansion.

\noindent \textbf{\textit{Task Diversity Collapse.}} A research question concerns the collapse of task diversity, which can arise when the system samples from its own high-probability regions without external correction~\cite{jia2025writingzerobridgegapnonverifiable}. As a result, the proposer tends to favor structurally clean and familiar tasks that align with the model’s pretraining distribution, while systematically neglecting low-frequency yet cognitively demanding patterns such as subtle metaphors or rare logical paradoxes~\cite{ICLR2025_be06e380}.

\noindent \textbf{\textit{Adaptive Curriculum Construction.}} Another research question is how to construct an automated curriculum signal for task proposal. Without an observable success metric, the proposer cannot reliably estimate the solver’s current zone of proximal development~\cite{ef4d7fb0-848f-3480-8634-d49a5f5c57df}. Consequently, generated tasks may oscillate between trivial prompts with limited improvement and overly difficult ones that produce noisy or vanishing learning signals. While difficulty in verifiable domains can be calibrated through success rates, it remains ill-defined in open domains.

\subsection{Challenge II: Reliable Evaluation and Long-Horizon Reward Signals}

In unverifiable domains, evaluation in self-play systems emerges as a central challenge. Unlike deterministic tasks that rely on exact matching or rule-based verification~\cite{Guo_2025,kimiteam2025kimik15scalingreinforcement}, these tasks are judged by subjective and inherently multi-dimensional criteria, such as coherence, creativity, utility, and impact~\cite{jia2025writingzerobridgegapnonverifiable}. This subsection examines two research questions.

\noindent \textbf{\textit{Reliable Subjective Evaluation.}} A critical gap lies in evaluating tasks such as creative writing or open-ended dialogue, which lack explicit ground truth. Existing approaches largely rely on scalar or generative reward models trained from human preference data under the reinforcement learning paradigm~\cite{Su2025CrossingTR,xu2025unifiedpairwiseframeworkrlhf}. However, the generalization ability of such reward models is constrained by the limited annotated preference data~\cite{zhong2025comprehensivesurveyrewardmodels}, \textbf{\textit{often encouraging the model to learn stylistic heuristics rather than semantic quality}}. Additionally, they are \textbf{\textit{vulnerable to reward hacking behaviors}}. The model tends to exploit superficial correlations between verbosity and higher reward signals~\cite{jia2025writingzerobridgegapnonverifiable}. We quantify this through experiments (see Table~\ref{tab:reward_hacking}). We annotate 150 samples from RubricBench~\cite{zhang2026rubricbenchaligningmodelgeneratedrubrics}, covering three types of reward hacking: length bias, formatting bias, and tone bias, with 50 samples for each type. We evaluate three groups of models: closed models, open models, and generative reward models, under w/o rubric and w/ rubric settings.




We have the following observations. (1) Among the three hacking types, most models achieve higher accuracy on tone bias. For example, GPT-OSS-120B achieves a preference accuracy of $0.94$ on tone bias. In contrast, performance on formatting bias is generally lower without rubric guidance, with Qwen3.5-Plus achieving $0.38$ and DeepSeek-V3.2 $0.28$. This suggests that higher apparent confidence or professional terminology is easier for models to capture, while surface formatting bias is more likely to mislead judgments. (2) Under the w/ rubric setting, open models achieve the best overall performance. In particular, GPT-OSS-120B and GPT-OSS-20B obtain the highest average accuracies of $0.84$ and $0.83$, respectively. This is consistent with their common use as reward models in self-play studies for unverifiable domains. By contrast, existing generative reward models still show room for improvement in defending against reward hacking. (3) Introducing rubrics significantly improves robustness to reward hacking across all models. For example, the overall accuracy of DeepSeek-V3.2 increases from $0.36$ to $0.75$, while GPT-4o-Mini improves from $0.37$ to $0.64$. These results indicate that explicit rubrics can constrain the judgment criteria of models and reduce their reliance on superficial heuristics.

\noindent \textbf{\textit{Long-Horizon Reward Design.}} The second research question concerns how to define reward signals in long-horizon tasks. Taking scientific discovery as an example, designing a reward function that meaningfully captures novelty, scientific impact, or reproducibility remains an unresolved problem~\cite{zhou2025autonomousagentsscientificdiscovery}. Recent studies on autonomous scientific discovery systems further highlight that scientific exploration involves sparse and delayed feedback, complex interdependencies across sub-tasks, and evaluation standards that may only become clear after extensive experimentation~\cite{feng2026internagent15unifiedagenticframework, internagentteam2025internagentagentscientist}. Additionally, studies on software development highlight that multi-stage workflows such as software design, implementation, and testing constitute a long-horizon pipeline, where strong interdependencies make it difficult to design intermediate reward signals~\cite{li-etal-2025-prompting}.


\subsection{Solution}

We propose two solutions for LLM self-play in unverifiable domains: the Generation-Evolution Pipeline for task generation and Causality-Aware Reward Enhancement for task evaluation.

\noindent \textbf{\textit{Generation-Evolution Pipeline.}} To address the challenges of diversity collapse and curriculum adaptation in task generation, we employ a Generation-Evolution Pipeline. In the Generation component, a novelty-seeking objective is introduced to penalize high-probability repetitions and encourage exploration of low-density regions in the model’s pretraining distribution. At the same time, a global historical pool of high-value trajectories is maintained to prevent catastrophic forgetting and stabilize the training distribution. The Evolution component manages the iterative updates of the self-play system through a feedback-driven loop, adjusting task difficulty. For example, when the solver’s success rate within a difficulty bin exceeds a threshold, the Evolution component triggers a complexity adjustment via prompt modification, achieving curriculum adaptation.



\noindent \textbf{\textit{Causality-Aware Reward Enhancement.}} To address the challenges of reliable evaluation and long-horizon reward signals in unverifiable tasks, we design a Causality-Aware Reward Enhancement mechanism. This method employs a process-based verifier to inspect the reasoning path and decomposes the reward model into key causal factors, which reflect true task quality, and non-causal spurious factors, such as length or style. For non-causal factors, specialized classifiers are applied to detect reward hacking types. For the key causal factors, implicit rewards are incorporated using expert demonstrations or environmental feedback~\cite{zhao2026selfdistilledreasoneronpolicyselfdistillation}, and multi-dimensional evaluation spaces combined with confidence-weighted calibration ensure the robustness and reliability of reward signals. Additionally, the introduction of scoring rubrics can further enhance the interpretability of the rewards~\cite{gunjal2025rubricsrewardsreinforcementlearning, huang2025reinforcementlearningrubricanchors}.


\section{Trustworthy Self-Play for Reliable Self-Evolution}
Self-play can generate effective experience while reducing reliance on human-curated tasks. However, it also creates trust and safety risks because the system learns from a closed feedback loop~\cite{wang-etal-2025-model,wang-etal-2024-boosting-llm}. For endogenous risks, noisy self-supervision signals, shortcut strategies, and unsafe reasoning patterns can enter the loop and grow over iterations~\cite{pmlr-v267-kazdan25a, dohmatob2025strong}. For exogenous risks, external data, tools, and adversaries can further perturb inputs and feedback, pushing the system toward unsafe behavior or lower robustness~\cite{ayzen2025cloak}. This section analyzes challenges from \textbf{\textit{endogenous and exogenous sources}} and proposes solutions to constrain the self-play process.

\subsection{Challenge I: Noisy Self-Supervision and Unsafe Cognitive Patterns}
Endogenous risks arise from the inherent dynamics within the self-play system, including self-supervised signals and reasoning patterns. In unverifiable tasks, solvers’ \textbf{\textit{unsupervised training relies on internal signals that are often unreliable, and solvers can also generate unsafe chain-of-thought reasoning without external supervision}}. This subsection discusses two essential research questions.

\noindent \textbf{\textit{Noisy Self-Supervision.}} In the math and open-world domains, the proposer can generate problems but may not provide reliably correct answers, making the solver’s training effectively unsupervised~\cite{huang2025rzeroselfevolvingreasoningllm,fang2025serl}. A key research question concerns the reliability of the solver’s self-generated signals are often noisy and susceptible to reward hacking or model collapse~\cite{fu2025selfverification, Fu2025theo}. For instance, majority voting used as a reward signal can reinforce consistent but incorrect answers~\cite{zuo2025ttrl}. Signals like self-consistency or entropy estimates are also often miscalibrated, and the loop can retain fluent, high-confidence errors while discarding low-confidence but informative samples~\cite{ICLR2025_93f25021}.


\noindent \textbf{\textit{Unsafe Cognitive Patterns.}} Although self-play reduces human intervention, the absence of supervision over intermediate reasoning steps can foster emergent cognitive patterns that are unsafe or undesirable. For example, LLMs may produce outputs such as: ``The aim is to outsmart all these groups of intelligent machines and less intelligent humans. This is for the brains behind the future,'' reflecting reasoning that could be misaligned with intended safe behavior~\cite{zhao2025absolutezeroreinforcedselfplay}. Such emergent reasoning tendencies may amplify through self-play iterations, as the solver reinforces risky patterns without external correction.~\cite{wang2024probingsafetyresponseboundary} also suggests that safety-aligned LLMs still retain latent unsafe decoding paths, which could be reinforced through self-play.

\subsection{Challenge II: Data Poisoning and Interface Manipulation}
In LLM-based self-play systems, the training loop interacts with multiple external interfaces, including data sources, feedback channels, evaluation tools, and retrieval systems. These interfaces introduce potential attack surfaces where malicious or corrupted signals may enter the pipeline. Unlike endogenous risks that emerge from internal learning dynamics, \textbf{\textit{exogenous risks arise when external actors intentionally inject harmful information into the system}}. This subsection highlights two research questions related to external threats.


\noindent \textbf{\textit{Data Poisoning.}} One research question concerns how poisoned or manipulated signals can infiltrate the training pipeline. Data poisoning may occur through compromised seed datasets, online synthesis pools, manipulated human feedback, tampered evaluation logs, or polluted retrieval corpora~\cite{li2025backdoorllm,shi2026prompt_injection_tool_selection}. In a self-play setting, such corruption can be amplified because poisoned samples may be replayed, reinforced by reward signals, and propagated across training iterations. Over time, these signals can steer the model toward attacker-chosen behaviors, even when the initial contamination is relatively small~\cite{li2025backdoorllm, 10.5555/3766078.3766275}.

\noindent \textbf{\textit{Interface Manipulation.}} Another research question concerns the vulnerability of external interfaces. Attackers may bias the reward pathway without directly modifying the underlying dataset. For example, they may manipulate the judge model, tool outputs, or scoring endpoints so that unsafe or incorrect generations receive high rewards, while safe behaviors are penalized~\cite{zhao2025one, tong2025badjudge}. In addition, memory systems can be exploited to inject carefully crafted misleading context, encouraging the model to rely on untrusted cues~\cite{10.5555/3766078.3766275}. 

\subsection{Solution}

We propose safety governance mechanisms: Behavioral Trustworthiness Regulation for endogenous risks and Provenance-Grounded Trust Certification for exogenous interface risks.


\noindent \textbf{\textit{Behavioral Trustworthiness Regulation.}} To mitigate noisy self-supervision, this component regulates behavior via three mechanisms. It integrates process supervision~\cite{lightman2024lets, zhao2025absolutezeroreinforcedselfplay} with outcome rewards to credit individual reasoning steps, penalizing correct answers derived from flawed paths. A calibration probe further tracks the gap between expressed confidence and empirical accuracy, flagging reasoning chains whose answers change under minor prompt rephrasing as shortcut-reliant. To mitigate unsafe cognitive patterns, the component applies a step-level, safety-aware KL penalty that targets only problematic steps, preserving the valid parts of the trajectory~\cite{NEURIPS2022_b1efde53, NEURIPS2024_bcfcf723}.




\noindent \textbf{\textit{Provenance-Grounded Trust Certification.}} To address exogenous risks, we propose Provenance-Grounded Trust Certification, which defines a trust boundary for all data entering the training pipeline. For data poisoning, this mechanism mandates verifiable origin metadata and applies anomaly scoring over provenance graphs to flag injection patterns that deviate from expected generation dynamics; once an anomaly is detected, the pipeline blocks the affected source and rolls back to a secure checkpoint. For interface manipulation, the component periodically re-scores random samples with the evaluator~\cite{NEURIPS2023_91f18a12, 10.1145/3701716.3715265} and withholds the batch when score divergence exceeds a preset margin, preventing policy corruption.

\section{Conclusion}


This paper argues that real-world LLM self-play systems should achieve effectiveness, generalizability, and trustworthiness under data scarcity. We outline three blue-sky directions: (1) data-centric self-play for effective experience generation; (2) generalizable self-play for open-world and unverifiable tasks; and (3) trustworthy self-play for reliable self-evolution. The paper summarizes challenges and potential solutions, aiming to provide a roadmap and promote a sustainable and trustworthy paradigm for LLM self-play.





\bibliographystyle{IEEEtran}
\bibliography{custom}

@misc{yang2025zeroguiautomatingonlinegui,
      title={ZeroGUI: Automating Online GUI Learning at Zero Human Cost}, 
      author={Chenyu Yang and Shiqian Su and Shi Liu and Xuan Dong and Yue Yu and Weijie Su and Xuehui Wang and Zhaoyang Liu and Jinguo Zhu and Hao Li and Wenhai Wang and Yu Qiao and Xizhou Zhu and Jifeng Dai},
      year={2025},
      eprint={2505.23762},
      archivePrefix={arXiv},
      primaryClass={cs.AI},
      url={https://arxiv.org/abs/2505.23762}, 
}

@misc{zhang2025pathselfevolvinglargelanguage,
      title={The Path of Self-Evolving Large Language Models: Achieving Data-Efficient Learning via Intrinsic Feedback}, 
      author={Hangfan Zhang and Siyuan Xu and Zhimeng Guo and Huaisheng Zhu and Shicheng Liu and Xinrun Wang and Qiaosheng Zhang and Yang Chen and Peng Ye and Lei Bai and Shuyue Hu},
      year={2025},
      eprint={2510.02752},
      archivePrefix={arXiv},
      primaryClass={cs.CL},
      url={https://arxiv.org/abs/2510.02752}, 
}

@misc{sun2025agenticselflearningllmssearch,
      title={Towards Agentic Self-Learning LLMs in Search Environment}, 
      author={Wangtao Sun and Xiang Cheng and Jialin Fan and Yao Xu and Xing Yu and Shizhu He and Jun Zhao and Kang Liu},
      year={2025},
      eprint={2510.14253},
      archivePrefix={arXiv},
      primaryClass={cs.AI},
      url={https://arxiv.org/abs/2510.14253}, 
}

@misc{feng2024extremelydataefficientgenerativellmbased,
      title={An Extremely Data-efficient and Generative LLM-based Reinforcement Learning Agent for Recommenders}, 
      author={Shuang Feng and Grace Feng},
      year={2024},
      eprint={2408.16032},
      archivePrefix={arXiv},
      primaryClass={cs.LG},
      url={https://arxiv.org/abs/2408.16032}, 
}

@misc{ma2025s2rteachingllmsselfverify,
      title={S$^2$R: Teaching LLMs to Self-verify and Self-correct via Reinforcement Learning}, 
      author={Ruotian Ma and Peisong Wang and Cheng Liu and Xingyan Liu and Jiaqi Chen and Bang Zhang and Xin Zhou and Nan Du and Jia Li},
      year={2025},
      eprint={2502.12853},
      archivePrefix={arXiv},
      primaryClass={cs.CL},
      url={https://arxiv.org/abs/2502.12853}, 
}

@misc{yang2025ssrzerosimpleselfrewardingreinforcement,
      title={SSR-Zero: Simple Self-Rewarding Reinforcement Learning for Machine Translation}, 
      author={Wenjie Yang and Mao Zheng and Mingyang Song and Zheng Li and Sitong Wang},
      year={2025},
      eprint={2505.16637},
      archivePrefix={arXiv},
      primaryClass={cs.CL},
      url={https://arxiv.org/abs/2505.16637}, 
}

@inproceedings{
fang2025serl,
title={Se{RL}: Self-play Reinforcement Learning for Large Language Models with Limited Data},
author={Wenkai Fang and Shunyu Liu and Yang Zhou and Kongcheng Zhang and Tongya Zheng and Kaixuan Chen and Mingli Song and Dacheng Tao},
booktitle={The Thirty-ninth Annual Conference on Neural Information Processing Systems},
year={2025},
url={https://openreview.net/forum?id=ZF93vyH9He}
}

@inproceedings{SPAG,
title={Self-playing Adversarial Language Game Enhances {LLM} Reasoning},
author={Pengyu Cheng and Tianhao Hu and Han Xu and Zhisong Zhang and Yong Dai and Lei Han and nan du and Xiaolong Li},
booktitle={The Thirty-eighth Annual Conference on Neural Information Processing Systems},
year={2024},
url={https://openreview.net/forum?id=oCGkSH7ys2}
}

@inproceedings{SPC,
title={{SPC}: Evolving Self-Play Critic via Adversarial Games for {LLM} Reasoning},
author={Jiaqi Chen and Bang Zhang and Ruotian Ma and Peisong Wang and Xiaodan Liang and Zhaopeng Tu and Xiaolong Li and Kwan-Yee K. Wong},
booktitle={The Thirty-ninth Annual Conference on Neural Information Processing Systems},
year={2025},
url={https://openreview.net/forum?id=JddJvNSiHk}
}

@misc{SPIRAL,
      title={SPIRAL: Self-Play on Zero-Sum Games Incentivizes Reasoning via Multi-Agent Multi-Turn Reinforcement Learning}, 
      author={Bo Liu and Leon Guertler and Simon Yu and Zichen Liu and Penghui Qi and Daniel Balcells and Mickel Liu and Cheston Tan and Weiyan Shi and Min Lin and Wee Sun Lee and Natasha Jaques},
      year={2025},
      eprint={2506.24119},
      archivePrefix={arXiv},
      primaryClass={cs.AI},
      url={https://arxiv.org/abs/2506.24119}, 
}

@misc{Vision-Zero,
      title={Vision-Zero: Scalable VLM Self-Improvement via Strategic Gamified Self-Play}, 
      author={Qinsi Wang and Bo Liu and Tianyi Zhou and Jing Shi and Yueqian Lin and Yiran Chen and Hai Helen Li and Kun Wan and Wentian Zhao},
      year={2025},
      eprint={2509.25541},
      archivePrefix={arXiv},
      primaryClass={cs.CV},
      url={https://arxiv.org/abs/2509.25541}, 
}

@misc{SPELL,
      title={SPELL: Self-Play Reinforcement Learning for evolving Long-Context Language Models}, 
      author={Ziyi Yang and Weizhou Shen and Ruijun Chen and Chenliang Li and Fanqi Wan and Ming Yan and Xiaojun Quan and Fei Huang},
      year={2025},
      eprint={2509.23863},
      archivePrefix={arXiv},
      primaryClass={cs.CL},
      url={https://arxiv.org/abs/2509.23863}, 
}

@misc{MAE,
      title={Multi-Agent Evolve: LLM Self-Improve through Co-evolution}, 
      author={Yixing Chen and Yiding Wang and Siqi Zhu and Haofei Yu and Tao Feng and Muhan Zhang and Mostofa Patwary and Jiaxuan You},
      year={2025},
      eprint={2510.23595},
      archivePrefix={arXiv},
      primaryClass={cs.AI},
      url={https://arxiv.org/abs/2510.23595}, 
}

@misc{PSV,
      title={Propose, Solve, Verify: Self-Play Through Formal Verification}, 
      author={Alex Wilf and Pranjal Aggarwal and Bryan Parno and Daniel Fried and Louis-Philippe Morency and Paul Pu Liang and Sean Welleck},
      year={2025},
      eprint={2512.18160},
      archivePrefix={arXiv},
      primaryClass={cs.AI},
      url={https://arxiv.org/abs/2512.18160}, 
}

@misc{Meta-RL-Crypto,
      title={Meta-Learning Reinforcement Learning for Crypto-Return Prediction}, 
      author={Junqiao Wang and Zhaoyang Guan and Guanyu Liu and Tianze Xia and Xianzhi Li and Shuo Yin and Xinyuan Song and Chuhan Cheng and Tianyu Shi and Alex Lee},
      year={2025},
      eprint={2509.09751},
      archivePrefix={arXiv},
      primaryClass={cs.LG},
      url={https://arxiv.org/abs/2509.09751}, 
}

@inproceedings{sukhbaatar2018intrinsic,
title={Intrinsic Motivation and Automatic Curricula via Asymmetric Self-Play},
author={Sainbayar Sukhbaatar and Zeming Lin and Ilya Kostrikov and Gabriel Synnaeve and Arthur Szlam and Rob Fergus},
booktitle={International Conference on Learning Representations},
year={2018},
url={https://openreview.net/forum?id=SkT5Yg-RZ},
}

@misc{openai2021asymmetricselfplayautomaticgoal,
      title={Asymmetric self-play for automatic goal discovery in robotic manipulation}, 
      author={OpenAI OpenAI and Matthias Plappert and Raul Sampedro and Tao Xu and Ilge Akkaya and Vineet Kosaraju and Peter Welinder and Ruben D'Sa and Arthur Petron and Henrique P. d. O. Pinto and Alex Paino and Hyeonwoo Noh and Lilian Weng and Qiming Yuan and Casey Chu and Wojciech Zaremba},
      year={2021},
      eprint={2101.04882},
      archivePrefix={arXiv},
      primaryClass={cs.LG},
      url={https://arxiv.org/abs/2101.04882}, 
}

@misc{chen2025selfquestioninglanguagemodels,
      title={Self-Questioning Language Models}, 
      author={Lili Chen and Mihir Prabhudesai and Katerina Fragkiadaki and Hao Liu and Deepak Pathak},
      year={2025},
      eprint={2508.03682},
      archivePrefix={arXiv},
      primaryClass={cs.LG},
      url={https://arxiv.org/abs/2508.03682}, 
}

@misc{wei2025learningposeproblemsreasoningdriven,
      title={Learning to Pose Problems: Reasoning-Driven and Solver-Adaptive Data Synthesis for Large Reasoning Models}, 
      author={Yongxian Wei and Yilin Zhao and Li Shen and Xinrui Chen and Runxi Cheng and Sinan Du and Hao Yu and Gang Liu and Jiahong Yan and Chun Yuan and Dian Li},
      year={2025},
      eprint={2511.09907},
      archivePrefix={arXiv},
      primaryClass={cs.AI},
      url={https://arxiv.org/abs/2511.09907}, 
}

@misc{he2025visplayselfevolvingvisionlanguagemodels,
      title={VisPlay: Self-Evolving Vision-Language Models from Images}, 
      author={Yicheng He and Chengsong Huang and Zongxia Li and Jiaxin Huang and Yonghui Yang},
      year={2025},
      eprint={2511.15661},
      archivePrefix={arXiv},
      primaryClass={cs.CV},
      url={https://arxiv.org/abs/2511.15661}, 
}

@inproceedings{wang-etal-2024-boosting-llm,
    title = "Boosting {LLM} Agents with Recursive Contemplation for Effective Deception Handling",
    author = "Wang, Shenzhi  and
      Liu, Chang  and
      Zheng, Zilong  and
      Qi, Siyuan  and
      Chen, Shuo  and
      Yang, Qisen  and
      Zhao, Andrew  and
      Wang, Chaofei  and
      Song, Shiji  and
      Huang, Gao",
    editor = "Ku, Lun-Wei  and
      Martins, Andre  and
      Srikumar, Vivek",
    booktitle = "Findings of the Association for Computational Linguistics: ACL 2024",
    month = aug,
    year = "2024",
    address = "Bangkok, Thailand",
    publisher = "Association for Computational Linguistics",
    url = "https://aclanthology.org/2024.findings-acl.591/",
    doi = "10.18653/v1/2024.findings-acl.591",
    pages = "9909--9953"
}

@inproceedings{wang-etal-2025-model,
    title = "Model Surgery: Modulating {LLM}{'}s Behavior Via Simple Parameter Editing",
    author = "Wang, Huanqian  and
      Yue, Yang  and
      Lu, Rui  and
      Shi, Jingxin  and
      Zhao, Andrew  and
      Wang, Shenzhi  and
      Song, Shiji  and
      Huang, Gao",
    editor = "Chiruzzo, Luis  and
      Ritter, Alan  and
      Wang, Lu",
    booktitle = "Proceedings of the 2025 Conference of the Nations of the Americas Chapter of the Association for Computational Linguistics: Human Language Technologies (Volume 1: Long Papers)",
    month = apr,
    year = "2025",
    address = "Albuquerque, New Mexico",
    publisher = "Association for Computational Linguistics",
    url = "https://aclanthology.org/2025.naacl-long.321/",
    doi = "10.18653/v1/2025.naacl-long.321",
    pages = "6337--6357",
    ISBN = "979-8-89176-189-6"
}

@misc{zhao2025absolutezeroreinforcedselfplay,
      title={Absolute Zero: Reinforced Self-play Reasoning with Zero Data}, 
      author={Andrew Zhao and Yiran Wu and Yang Yue and Tong Wu and Quentin Xu and Yang Yue and Matthieu Lin and Shenzhi Wang and Qingyun Wu and Zilong Zheng and Gao Huang},
      year={2025},
      eprint={2505.03335},
      archivePrefix={arXiv},
      primaryClass={cs.LG},
      url={https://arxiv.org/abs/2505.03335}, 
}

@misc{huang2025rzeroselfevolvingreasoningllm,
      title={R-Zero: Self-Evolving Reasoning LLM from Zero Data}, 
      author={Chengsong Huang and Wenhao Yu and Xiaoyang Wang and Hongming Zhang and Zongxia Li and Ruosen Li and Jiaxin Huang and Haitao Mi and Dong Yu},
      year={2025},
      eprint={2508.05004},
      archivePrefix={arXiv},
      primaryClass={cs.LG},
      url={https://arxiv.org/abs/2508.05004}, 
}

@misc{cui2025processreinforcementimplicitrewards,
      title={Process Reinforcement through Implicit Rewards}, 
      author={Ganqu Cui and Lifan Yuan and Zefan Wang and Hanbin Wang and Yuchen Zhang and Jiacheng Chen and Wendi Li and Bingxiang He and Yuchen Fan and Tianyu Yu and Qixin Xu and Weize Chen and Jiarui Yuan and Huayu Chen and Kaiyan Zhang and Xingtai Lv and Shuo Wang and Yuan Yao and Xu Han and Hao Peng and Yu Cheng and Zhiyuan Liu and Maosong Sun and Bowen Zhou and Ning Ding},
      year={2025},
      eprint={2502.01456},
      archivePrefix={arXiv},
      primaryClass={cs.LG},
      url={https://arxiv.org/abs/2502.01456}, 
}

@misc{guo2024deepseekcoderlargelanguagemodel,
      title={DeepSeek-Coder: When the Large Language Model Meets Programming -- The Rise of Code Intelligence}, 
      author={Daya Guo and Qihao Zhu and Dejian Yang and Zhenda Xie and Kai Dong and Wentao Zhang and Guanting Chen and Xiao Bi and Y. Wu and Y. K. Li and Fuli Luo and Yingfei Xiong and Wenfeng Liang},
      year={2024},
      eprint={2401.14196},
      archivePrefix={arXiv},
      primaryClass={cs.SE},
      url={https://arxiv.org/abs/2401.14196}, 
}

@inproceedings{li-etal-2025-prompting,
    title = "Prompting Large Language Models to Tackle the Full Software Development Lifecycle: A Case Study",
    author = "Li, Bowen  and
      Wu, Wenhan  and
      Tang, Ziwei  and
      Shi, Lin  and
      Yang, John  and
      Li, Jinyang  and
      Yao, Shunyu  and
      Qian, Chen  and
      Hui, Binyuan  and
      Zhang, Qicheng  and
      Yu, Zhiyin  and
      Du, He  and
      Yang, Ping  and
      Lin, Dahua  and
      Peng, Chao  and
      Chen, Kai",
    editor = "Rambow, Owen  and
      Wanner, Leo  and
      Apidianaki, Marianna  and
      Al-Khalifa, Hend  and
      Eugenio, Barbara Di  and
      Schockaert, Steven",
    booktitle = "Proceedings of the 31st International Conference on Computational Linguistics",
    month = jan,
    year = "2025",
    address = "Abu Dhabi, UAE",
    publisher = "Association for Computational Linguistics",
    url = "https://aclanthology.org/2025.coling-main.502/",
    pages = "7511--7531"
}

@misc{internagentteam2025internagentagentscientist,
      title={InternAgent: When Agent Becomes the Scientist -- Building Closed-Loop System from Hypothesis to Verification}, 
      author={InternAgent Team and Bo Zhang and Shiyang Feng and Xiangchao Yan and Jiakang Yuan and Runmin Ma and Yusong Hu and Zhiyin Yu and Xiaohan He and Songtao Huang and Shaowei Hou and Zheng Nie and Zhilong Wang and Jinyao Liu and Tianshuo Peng and Peng Ye and Dongzhan Zhou and Shufei Zhang and Xiaosong Wang and Yilan Zhang and Meng Li and Zhongying Tu and Xiangyu Yue and Wangli Ouyang and Bowen Zhou and Lei Bai},
      year={2025},
      eprint={2505.16938},
      archivePrefix={arXiv},
      primaryClass={cs.AI},
      url={https://arxiv.org/abs/2505.16938}, 
}

@article{jones2024ai,
  title={The AI revolution is running out of data},
  author={Jones, N},
  journal={What can researchers do},
  year={2024}
}

@inproceedings{
zuo2025ttrl,
title={{TTRL}: Test-Time Reinforcement Learning},
author={Yuxin Zuo and Kaiyan Zhang and Li Sheng and Shang Qu and Ganqu Cui and Xuekai Zhu and Haozhan Li and Yuchen Zhang and Xinwei Long and Ermo Hua and Biqing Qi and Youbang Sun and Zhiyuan Ma and Lifan Yuan and Ning Ding and Bowen Zhou},
booktitle={The Thirty-ninth Annual Conference on Neural Information Processing Systems},
year={2025},
url={https://openreview.net/forum?id=VuVhgEiu20}
}

@techreport{silver2025era,
  title={Welcome to the Era of Experience},
  author={Silver, David and Sutton, Richard S.},
  institution={Google AI},
  year={2025}
}

@misc{kaplan2020scalinglawsneurallanguage,
      title={Scaling Laws for Neural Language Models}, 
      author={Jared Kaplan and Sam McCandlish and Tom Henighan and Tom B. Brown and Benjamin Chess and Rewon Child and Scott Gray and Alec Radford and Jeffrey Wu and Dario Amodei},
      year={2020},
      eprint={2001.08361},
      archivePrefix={arXiv},
      primaryClass={cs.LG},
      url={https://arxiv.org/abs/2001.08361}, 
}

@article{silver2017mastering,
  title={Mastering the game of go without human knowledge},
  author={Silver, David and Schrittwieser, Julian and Simonyan, Karen and Antonoglou, Ioannis and Huang, Aja and Guez, Arthur and Hubert, Thomas and Baker, Lucas and Lai, Matthew and Bolton, Adrian and others},
  journal={nature},
  volume={550},
  number={7676},
  pages={354--359},
  year={2017},
  publisher={Nature Publishing Group UK London}
}

@inproceedings{
anonymous2026semantic,
title={Semantic Voting: A Self-Evaluation-Free Approach for Efficient {LLM} Self-Improvement on Unverifiable Open-ended Tasks},
author={Anonymous},
booktitle={The Fourteenth International Conference on Learning Representations},
year={2026},
url={https://openreview.net/forum?id=7AlPbFkcs3}
}

@misc{jia2025writingzerobridgegapnonverifiable,
      title={Writing-Zero: Bridge the Gap Between Non-verifiable Tasks and Verifiable Rewards}, 
      author={Ruipeng Jia and Yunyi Yang and Yongbo Gai and Kai Luo and Shihao Huang and Jianhe Lin and Xiaoxi Jiang and Guanjun Jiang},
      year={2025},
      eprint={2506.00103},
      archivePrefix={arXiv},
      primaryClass={cs.CL},
      url={https://arxiv.org/abs/2506.00103}, 
}

@misc{zhong2025comprehensivesurveyrewardmodels,
      title={A Comprehensive Survey of Reward Models: Taxonomy, Applications, Challenges, and Future}, 
      author={Jialun Zhong and Wei Shen and Yanzeng Li and Songyang Gao and Hua Lu and Yicheng Chen and Yang Zhang and Wei Zhou and Jinjie Gu and Lei Zou},
      year={2025},
      eprint={2504.12328},
      archivePrefix={arXiv},
      primaryClass={cs.CL},
      url={https://arxiv.org/abs/2504.12328}, 
}

@article{Guo_2025,
   title={DeepSeek-R1 incentivizes reasoning in LLMs through reinforcement learning},
   volume={645},
   ISSN={1476-4687},
   url={http://dx.doi.org/10.1038/s41586-025-09422-z},
   DOI={10.1038/s41586-025-09422-z},
   number={8081},
   journal={Nature},
   publisher={Springer Science and Business Media LLC},
   author={Guo, Daya and Yang, Dejian and Zhang, Haowei and Song, Junxiao and Wang, Peiyi and Zhu, Qihao and Xu, Runxin and Zhang, Ruoyu and Ma, Shirong and Bi, Xiao and Zhang, Xiaokang and Yu, Xingkai and Wu, Yu and Wu, Z. F. and Gou, Zhibin and Shao, Zhihong and Li, Zhuoshu and Gao, Ziyi and Liu, Aixin and Xue, Bing and Wang, Bingxuan and Wu, Bochao and Feng, Bei and Lu, Chengda and Zhao, Chenggang and Deng, Chengqi and Ruan, Chong and Dai, Damai and Chen, Deli and Ji, Dongjie and Li, Erhang and Lin, Fangyun and Dai, Fucong and Luo, Fuli and Hao, Guangbo and Chen, Guanting and Li, Guowei and Zhang, H. and Xu, Hanwei and Ding, Honghui and Gao, Huazuo and Qu, Hui and Li, Hui and Guo, Jianzhong and Li, Jiashi and Chen, Jingchang and Yuan, Jingyang and Tu, Jinhao and Qiu, Junjie and Li, Junlong and Cai, J. L. and Ni, Jiaqi and Liang, Jian and Chen, Jin and Dong, Kai and Hu, Kai and You, Kaichao and Gao, Kaige and Guan, Kang and Huang, Kexin and Yu, Kuai and Wang, Lean and Zhang, Lecong and Zhao, Liang and Wang, Litong and Zhang, Liyue and Xu, Lei and Xia, Leyi and Zhang, Mingchuan and Zhang, Minghua and Tang, Minghui and Zhou, Mingxu and Li, Meng and Wang, Miaojun and Li, Mingming and Tian, Ning and Huang, Panpan and Zhang, Peng and Wang, Qiancheng and Chen, Qinyu and Du, Qiushi and Ge, Ruiqi and Zhang, Ruisong and Pan, Ruizhe and Wang, Runji and Chen, R. J. and Jin, R. L. and Chen, Ruyi and Lu, Shanghao and Zhou, Shangyan and Chen, Shanhuang and Ye, Shengfeng and Wang, Shiyu and Yu, Shuiping and Zhou, Shunfeng and Pan, Shuting and Li, S. S. and Zhou, Shuang and Wu, Shaoqing and Yun, Tao and Pei, Tian and Sun, Tianyu and Wang, T. and Zeng, Wangding and Liu, Wen and Liang, Wenfeng and Gao, Wenjun and Yu, Wenqin and Zhang, Wentao and Xiao, W. L. and An, Wei and Liu, Xiaodong and Wang, Xiaohan and Chen, Xiaokang and Nie, Xiaotao and Cheng, Xin and Liu, Xin and Xie, Xin and Liu, Xingchao and Yang, Xinyu and Li, Xinyuan and Su, Xuecheng and Lin, Xuheng and Li, X. Q. and Jin, Xiangyue and Shen, Xiaojin and Chen, Xiaosha and Sun, Xiaowen and Wang, Xiaoxiang and Song, Xinnan and Zhou, Xinyi and Wang, Xianzu and Shan, Xinxia and Li, Y. K. and Wang, Y. Q. and Wei, Y. X. and Zhang, Yang and Xu, Yanhong and Li, Yao and Zhao, Yao and Sun, Yaofeng and Wang, Yaohui and Yu, Yi and Zhang, Yichao and Shi, Yifan and Xiong, Yiliang and He, Ying and Piao, Yishi and Wang, Yisong and Tan, Yixuan and Ma, Yiyang and Liu, Yiyuan and Guo, Yongqiang and Ou, Yuan and Wang, Yuduan and Gong, Yue and Zou, Yuheng and He, Yujia and Xiong, Yunfan and Luo, Yuxiang and You, Yuxiang and Liu, Yuxuan and Zhou, Yuyang and Zhu, Y. X. and Huang, Yanping and Li, Yaohui and Zheng, Yi and Zhu, Yuchen and Ma, Yunxian and Tang, Ying and Zha, Yukun and Yan, Yuting and Ren, Z. Z. and Ren, Zehui and Sha, Zhangli and Fu, Zhe and Xu, Zhean and Xie, Zhenda and Zhang, Zhengyan and Hao, Zhewen and Ma, Zhicheng and Yan, Zhigang and Wu, Zhiyu and Gu, Zihui and Zhu, Zijia and Liu, Zijun and Li, Zilin and Xie, Ziwei and Song, Ziyang and Pan, Zizheng and Huang, Zhen and Xu, Zhipeng and Zhang, Zhongyu and Zhang, Zhen},
   year={2025},
   month=sep, pages={633–638} }

@misc{kimiteam2025kimik15scalingreinforcement,
      title={Kimi k1.5: Scaling Reinforcement Learning with LLMs}, 
      author={Kimi Team and Angang Du and Bofei Gao and Bowei Xing and Changjiu Jiang and Cheng Chen and Cheng Li and Chenjun Xiao and Chenzhuang Du and Chonghua Liao and Chuning Tang and Congcong Wang and Dehao Zhang and Enming Yuan and Enzhe Lu and Fengxiang Tang and Flood Sung and Guangda Wei and Guokun Lai and Haiqing Guo and Han Zhu and Hao Ding and Hao Hu and Hao Yang and Hao Zhang and Haotian Yao and Haotian Zhao and Haoyu Lu and Haoze Li and Haozhen Yu and Hongcheng Gao and Huabin Zheng and Huan Yuan and Jia Chen and Jianhang Guo and Jianlin Su and Jianzhou Wang and Jie Zhao and Jin Zhang and Jingyuan Liu and Junjie Yan and Junyan Wu and Lidong Shi and Ling Ye and Longhui Yu and Mengnan Dong and Neo Zhang and Ningchen Ma and Qiwei Pan and Qucheng Gong and Shaowei Liu and Shengling Ma and Shupeng Wei and Sihan Cao and Siying Huang and Tao Jiang and Weihao Gao and Weimin Xiong and Weiran He and Weixiao Huang and Weixin Xu and Wenhao Wu and Wenyang He and Xianghui Wei and Xianqing Jia and Xingzhe Wu and Xinran Xu and Xinxing Zu and Xinyu Zhou and Xuehai Pan and Y. Charles and Yang Li and Yangyang Hu and Yangyang Liu and Yanru Chen and Yejie Wang and Yibo Liu and Yidao Qin and Yifeng Liu and Ying Yang and Yiping Bao and Yulun Du and Yuxin Wu and Yuzhi Wang and Zaida Zhou and Zhaoji Wang and Zhaowei Li and Zhen Zhu and Zheng Zhang and Zhexu Wang and Zhilin Yang and Zhiqi Huang and Zihao Huang and Ziyao Xu and Zonghan Yang and Zongyu Lin},
      year={2025},
      eprint={2501.12599},
      archivePrefix={arXiv},
      primaryClass={cs.AI},
      url={https://arxiv.org/abs/2501.12599}, 
}

@misc{zhou2025autonomousagentsscientificdiscovery,
      title={Autonomous Agents for Scientific Discovery: Orchestrating Scientists, Language, Code, and Physics}, 
      author={Lianhao Zhou and Hongyi Ling and Cong Fu and Yepeng Huang and Michael Sun and Wendi Yu and Xiaoxuan Wang and Xiner Li and Xingyu Su and Junkai Zhang and Xiusi Chen and Chenxing Liang and Xiaofeng Qian and Heng Ji and Wei Wang and Marinka Zitnik and Shuiwang Ji},
      year={2025},
      eprint={2510.09901},
      archivePrefix={arXiv},
      primaryClass={cs.AI},
      url={https://arxiv.org/abs/2510.09901}, 
}

@inproceedings{
song2025mind,
title={Mind the Gap: Examining the Self-Improvement Capabilities of Large Language Models},
author={Yuda Song and Hanlin Zhang and Carson Eisenach and Sham M. Kakade and Dean Foster and Udaya Ghai},
booktitle={The Thirteenth International Conference on Learning Representations},
year={2025},
url={https://openreview.net/forum?id=mtJSMcF3ek}
}

@InProceedings{Chen2024SPIN,
  title = 	 {Self-Play Fine-Tuning Converts Weak Language Models to Strong Language Models},
  author =       {Chen, Zixiang and Deng, Yihe and Yuan, Huizhuo and Ji, Kaixuan and Gu, Quanquan},
  booktitle = 	 {Proceedings of the 41st International Conference on Machine Learning},
  pages = 	 {6621--6642},
  year = 	 {2024},
  editor = 	 {Salakhutdinov, Ruslan and Kolter, Zico and Heller, Katherine and Weller, Adrian and Oliver, Nuria and Scarlett, Jonathan and Berkenkamp, Felix},
  volume = 	 {235},
  series = 	 {Proceedings of Machine Learning Research},
  month = 	 {21--27 Jul},
  publisher =    {PMLR},
  url = 	 {https://proceedings.mlr.press/v235/chen24j.html},
}

@inproceedings{
yue2025does,
title={Does Reinforcement Learning Really Incentivize Reasoning Capacity in {LLM}s Beyond the Base Model?},
author={Yang Yue and Zhiqi Chen and Rui Lu and Andrew Zhao and Zhaokai Wang and Yang Yue and Shiji Song and Gao Huang},
booktitle={2nd AI for Math Workshop @ ICML 2025},
year={2025},
url={https://openreview.net/forum?id=upehLVgq1b}
}

@inproceedings{dohmatob2025strong,
 author = {Dohmatob, Elvis and Feng, Yunzhen and Subramonian, Arjun and Kempe, Julia},
 booktitle = {International Conference on Learning Representations},
 editor = {Y. Yue and A. Garg and N. Peng and F. Sha and R. Yu},
 pages = {15656--15691},
 title = {Strong Model Collapse},
 url = {https://proceedings.iclr.cc/paper_files/paper/2025/file/284afdc2309f9667d2d4fb9290235b0c-Paper-Conference.pdf},
 volume = {2025},
 year = {2025}
}

@inproceedings{Fu2025theo,
 author = {Fu, Shi and Wang, Yingjie and Chen, Yuzhu and Tian, Xinmei and Tao, Dacheng},
 booktitle = {International Conference on Learning Representations},
 editor = {Y. Yue and A. Garg and N. Peng and F. Sha and R. Yu},
 pages = {93254--93287},
 title = {A Theoretical Perspective: How to Prevent Model Collapse in Self-consuming Training Loops},
 url = {https://proceedings.iclr.cc/paper_files/paper/2025/file/e92e6d20f7b5db5c2a1b23964d170fe2-Paper-Conference.pdf},
 volume = {2025},
 year = {2025}
}

@inproceedings{shen-etal-2025-cycle,
    title = "{CYCLE}-{INSTRUCT}: Fully Seed-Free Instruction Tuning via Dual Self-Training and Cycle Consistency",
    author = "Shen, Zhanming  and
      Chen, Hao  and
      Tang, Yulei  and
      Zhu, Shaolin  and
      Ye, Wentao  and
      Hu, Xiaomeng  and
      Wang, Haobo  and
      Chen, Gang  and
      Zhao, Junbo",
    editor = "Christodoulopoulos, Christos  and
      Chakraborty, Tanmoy  and
      Rose, Carolyn  and
      Peng, Violet",
    booktitle = "Proceedings of the 2025 Conference on Empirical Methods in Natural Language Processing",
    month = nov,
    year = "2025",
    address = "Suzhou, China",
    publisher = "Association for Computational Linguistics",
    url = "https://aclanthology.org/2025.emnlp-main.258/",
    doi = "10.18653/v1/2025.emnlp-main.258",
    pages = "5123--5137",
    ISBN = "979-8-89176-332-6"
}

@inproceedings{ICLR2025_93f25021,
 author = {wen, xueru and Lou, Jie and Lu, Yaojie and Lin, Hongyu and XingYu and Lu, Xinyu and He, Ben and Han, Xianpei and Zhang, Debing and Sun, Le},
 booktitle = {International Conference on Learning Representations},
 editor = {Y. Yue and A. Garg and N. Peng and F. Sha and R. Yu},
 pages = {58769--58790},
 title = {Rethinking Reward Model Evaluation: Are We Barking up the Wrong Tree?},
 url = {https://proceedings.iclr.cc/paper_files/paper/2025/file/93f250215e4889119807b6fac3a57aec-Paper-Conference.pdf},
 volume = {2025},
 year = {2025}
}

@inproceedings{wang-etal-2025-removing,
    title = "Removing Prompt-template Bias in Reinforcement Learning from Human Feedback",
    author = "Wang, Chaojie  and
      Shi, Haonan  and
      Tian, Long  and
      An, Bo  and
      Yan, Shuicheng",
    editor = "Che, Wanxiang  and
      Nabende, Joyce  and
      Shutova, Ekaterina  and
      Pilehvar, Mohammad Taher",
    booktitle = "Findings of the Association for Computational Linguistics: ACL 2025",
    month = jul,
    year = "2025",
    address = "Vienna, Austria",
    publisher = "Association for Computational Linguistics",
    url = "https://aclanthology.org/2025.findings-acl.1237/",
    doi = "10.18653/v1/2025.findings-acl.1237",
    pages = "24110--24122",
    ISBN = "979-8-89176-256-5",
}

@InProceedings{pmlr-v267-kazdan25a,
  title = 	 {Collapse or Thrive: Perils and Promises of Synthetic Data in a Self-Generating World},
  author =       {Kazdan, Joshua and Schaeffer, Rylan and Dey, Apratim and Gerstgrasser, Matthias and Rafailov, Rafael and Donoho, David L. and Koyejo, Sanmi},
  booktitle = 	 {Proceedings of the 42nd International Conference on Machine Learning},
  pages = 	 {29469--29494},
  year = 	 {2025},
  editor = 	 {Singh, Aarti and Fazel, Maryam and Hsu, Daniel and Lacoste-Julien, Simon and Berkenkamp, Felix and Maharaj, Tegan and Wagstaff, Kiri and Zhu, Jerry},
  volume = 	 {267},
  series = 	 {Proceedings of Machine Learning Research},
  month = 	 {13--19 Jul},
  publisher =    {PMLR},
  url = 	 {https://proceedings.mlr.press/v267/kazdan25a.html},
}

@inproceedings{
fu2025selfverification,
title={Self-Verification Provably Prevents Model Collapse in Recursive Synthetic Training},
author={Shi Fu and Yingjie Wang and Yuzhu Chen and Li Shen and Dacheng Tao},
booktitle={The Thirty-ninth Annual Conference on Neural Information Processing Systems},
year={2025},
url={https://openreview.net/forum?id=X5Hk8aMs6w}
}

@inproceedings{ayzen2025cloak,
author = {Ayzenshteyn, Daniel and Weiss, Roy and Mirsky, Yisroel},
title = {Cloak, honey, trap: proactive defenses against LLM agents},
year = {2025},
isbn = {978-1-939133-52-6},
publisher = {USENIX Association},
address = {USA},
booktitle = {Proceedings of the 34th USENIX Conference on Security Symposium},
articleno = {415},
numpages = {20},
location = {Seattle, WA, USA},
series = {SEC '25}
}

@article{trinh2024solving,
  title={Solving olympiad geometry without human demonstrations},
  author={Trinh, Trieu H and Wu, Yuhuai and Le, Quoc V and He, He and Luong, Thang},
  journal={Nature},
  volume={625},
  number={7995},
  pages={476--482},
  year={2024},
  publisher={Nature Publishing Group UK London},
  url={https://www.nature.com/articles/s41586-023-06747-5}
}

@inproceedings{wang-etal-2023-self-instruct,
    title = "Self-Instruct: Aligning Language Models with Self-Generated Instructions",
    author = "Wang, Yizhong  and
      Kordi, Yeganeh  and
      Mishra, Swaroop  and
      Liu, Alisa  and
      Smith, Noah A.  and
      Khashabi, Daniel  and
      Hajishirzi, Hannaneh",
    editor = "Rogers, Anna  and
      Boyd-Graber, Jordan  and
      Okazaki, Naoaki",
    booktitle = "Proceedings of the 61st Annual Meeting of the Association for Computational Linguistics (Volume 1: Long Papers)",
    month = jul,
    year = "2023",
    address = "Toronto, Canada",
    publisher = "Association for Computational Linguistics",
    url = "https://aclanthology.org/2023.acl-long.754/",
    doi = "10.18653/v1/2023.acl-long.754",
    pages = "13484--13508"
}

@inproceedings{shi2026prompt_injection_tool_selection,
  title     = {Prompt Injection Attack to Tool Selection in LLM Agents},
  author    = {Shi, Jiawen and Yuan, Zenghui and Tie, Guiyao and Zhou, Pan and Gong, Neil Zhenqiang and Sun, Lichao},
  booktitle = {Proceedings of the Network and Distributed System Security Symposium },
  year      = {2026},
  month     = feb,
  address   = {San Diego, CA, USA},
  url       = {https://www.ndss-symposium.org/wp-content/uploads/2026-s675-paper.pdf}
}

@inproceedings{
li2025backdoorllm,
title={Backdoor{LLM}: A Comprehensive Benchmark for Backdoor Attacks and Defenses on Large Language Models},
author={Yige Li and Hanxun Huang and Yunhan Zhao and Xingjun Ma and Jun Sun},
booktitle={The Thirty-ninth Annual Conference on Neural Information Processing Systems Datasets and Benchmarks Track},
year={2025},
url={https://openreview.net/forum?id=sYLiY87mNn}
}

@inproceedings{
tong2025badjudge,
title={BadJudge: Backdoor Vulnerabilities of {LLM}-As-A-Judge},
author={Terry Tong and Fei Wang and Zhe Zhao and Muhao Chen},
booktitle={The Thirteenth International Conference on Learning Representations},
year={2025},
url={https://openreview.net/forum?id=eC2a2IndIt}
}

@inproceedings{
zhao2025one,
title={One Token to Fool {LLM}-as-a-Judge},
author={Yulai Zhao and Haolin Liu and Dian Yu and Sunyuan Kung and MEIJIA CHEN and Haitao Mi and Dong Yu},
booktitle={The 5th Workshop on Mathematical Reasoning and AI at NeurIPS 2025},
year={2025},
url={https://openreview.net/forum?id=M51kR6HXo4}
}

@inproceedings{10.5555/3766078.3766275,
author = {Zou, Wei and Geng, Runpeng and Wang, Binghui and Jia, Jinyuan},
title = {PoisonedRAG: knowledge corruption attacks to retrieval-augmented generation of large language models},
year = {2025},
isbn = {978-1-939133-52-6},
publisher = {USENIX Association},
address = {USA},
booktitle = {Proceedings of the 34th USENIX Conference on Security Symposium},
articleno = {197},
numpages = {18},
location = {Seattle, WA, USA},
series = {SEC '25}
}

@inproceedings{ICLR2024_82eec786,
 author = {Xu, Can and Sun, Qingfeng and Zheng, Kai and Geng, Xiubo and Zhao, Pu and Feng, Jiazhan and Tao, Chongyang and Lin, Qingwei and Jiang, Daxin},
 booktitle = {International Conference on Learning Representations},
 editor = {B. Kim and Y. Yue and S. Chaudhuri and K. Fragkiadaki and M. Khan and Y. Sun},
 pages = {30745--30766},
 title = {WizardLM: Empowering Large Pre-Trained Language Models to Follow Complex Instructions},
 url = {https://proceedings.iclr.cc/paper_files/paper/2024/file/82eec786fdfbbfa53450c5feb7d1ac92-Paper-Conference.pdf},
 volume = {2024},
 year = {2024}
}

@inproceedings{ICLR2025_be06e380,
 author = {Xu, Zhangchen and Jiang, Fengqing and Niu, Luyao and Deng, Yuntian and Poovendran, Radha and Choi, Yejin and Lin, Bill Yuchen},
 booktitle = {International Conference on Learning Representations},
 editor = {Y. Yue and A. Garg and N. Peng and F. Sha and R. Yu},
 pages = {76346--76382},
 title = {Magpie: Alignment Data Synthesis from Scratch by Prompting Aligned LLMs with Nothing},
 url = {https://proceedings.iclr.cc/paper_files/paper/2025/file/be06e3802e9411381feece79b4d960c1-Paper-Conference.pdf},
 volume = {2025},
 year = {2025}
}

@book{ef4d7fb0-848f-3480-8634-d49a5f5c57df,
 ISBN = {9780674576285},
 URL = {http://www.jstor.org/stable/j.ctvjf9vz4},
 author = {L. S. VYGOTSKY},
 publisher = {Harvard University Press},
 title = {Mind in Society: Development of Higher Psychological Processes},
 urldate = {2025-11-04},
 year = {1978}
}

@article{Su2025CrossingTR,
  title={Crossing the Reward Bridge: Expanding RL with Verifiable Rewards Across Diverse Domains},
  author={Yi Su and Dian Yu and Linfeng Song and Juntao Li and Haitao Mi and Zhaopeng Tu and Min Zhang and Dong Yu},
  journal={ArXiv},
  year={2025},
  volume={abs/2503.23829},
  url={https://api.semanticscholar.org/CorpusID:277451641}
}

@misc{xu2025unifiedpairwiseframeworkrlhf,
      title={A Unified Pairwise Framework for RLHF: Bridging Generative Reward Modeling and Policy Optimization}, 
      author={Wenyuan Xu and Xiaochen Zuo and Chao Xin and Yu Yue and Lin Yan and Yonghui Wu},
      year={2025},
      eprint={2504.04950},
      archivePrefix={arXiv},
      primaryClass={cs.LG},
      url={https://arxiv.org/abs/2504.04950}, 
}

@misc{zhao2026selfdistilledreasoneronpolicyselfdistillation,
      title={Self-Distilled Reasoner: On-Policy Self-Distillation for Large Language Models}, 
      author={Siyan Zhao and Zhihui Xie and Mengchen Liu and Jing Huang and Guan Pang and Feiyu Chen and Aditya Grover},
      year={2026},
      eprint={2601.18734},
      archivePrefix={arXiv},
      primaryClass={cs.LG},
      url={https://arxiv.org/abs/2601.18734}, 
}

@misc{liu2026selfplayevolvesselfsyntheticpipeline,
      title={Self-Play Only Evolves When Self-Synthetic Pipeline Ensures Learnable Information Gain}, 
      author={Wei Liu and Siya Qi and Yali Du and Yulan He},
      year={2026},
      eprint={2603.02218},
      archivePrefix={arXiv},
      primaryClass={cs.LG},
      url={https://arxiv.org/abs/2603.02218}, 
}

@misc{yang2025spendwiselymaximizingposttraining,
      title={Spend Wisely: Maximizing Post-Training Gains in Iterative Synthetic Data Bootstrapping}, 
      author={Pu Yang and Yunzhen Feng and Ziyuan Chen and Yuhang Wu and Zhuoyuan Li},
      year={2025},
      eprint={2501.18962},
      archivePrefix={arXiv},
      primaryClass={cs.LG},
      url={https://arxiv.org/abs/2501.18962}, 
}

@inproceedings{NEURIPS2019_fa7cdfad,
 author = {Rolnick, David and Ahuja, Arun and Schwarz, Jonathan and Lillicrap, Timothy and Wayne, Gregory},
 booktitle = {Advances in Neural Information Processing Systems},
 editor = {H. Wallach and H. Larochelle and A. Beygelzimer and F. d\textquotesingle Alch\'{e}-Buc and E. Fox and R. Garnett},
 pages = {},
 publisher = {Curran Associates, Inc.},
 title = {Experience Replay for Continual Learning},
 url = {https://proceedings.neurips.cc/paper_files/paper/2019/file/fa7cdfad1a5aaf8370ebeda47a1ff1c3-Paper.pdf},
 volume = {32},
 year = {2019}
}

@inproceedings{
lightman2024lets,
title={Let's Verify Step by Step},
author={Hunter Lightman and Vineet Kosaraju and Yuri Burda and Harrison Edwards and Bowen Baker and Teddy Lee and Jan Leike and John Schulman and Ilya Sutskever and Karl Cobbe},
booktitle={The Twelfth International Conference on Learning Representations},
year={2024},
url={https://openreview.net/forum?id=v8L0pN6EOi}
}

@inproceedings{NEURIPS2022_b1efde53,
 author = {Ouyang, Long and Wu, Jeffrey and Jiang, Xu and Almeida, Diogo and Wainwright, Carroll and Mishkin, Pamela and Zhang, Chong and Agarwal, Sandhini and Slama, Katarina and Ray, Alex and Schulman, John and Hilton, Jacob and Kelton, Fraser and Miller, Luke and Simens, Maddie and Askell, Amanda and Welinder, Peter and Christiano, Paul F and Leike, Jan and Lowe, Ryan},
 booktitle = {Advances in Neural Information Processing Systems},
 editor = {S. Koyejo and S. Mohamed and A. Agarwal and D. Belgrave and K. Cho and A. Oh},
 pages = {27730--27744},
 publisher = {Curran Associates, Inc.},
 title = {Training language models to follow instructions with human feedback},
 url = {https://proceedings.neurips.cc/paper_files/paper/2022/file/b1efde53be364a73914f58805a001731-Paper-Conference.pdf},
 volume = {35},
 year = {2022}
}

@inproceedings{NEURIPS2024_bcfcf723,
 author = {Wachi, Akifumi and Tran, Thien Q. and Sato, Rei and Tanabe, Takumi and Akimoto, Youhei},
 booktitle = {Advances in Neural Information Processing Systems},
 doi = {10.52202/079017-3319},
 editor = {A. Globerson and L. Mackey and D. Belgrave and A. Fan and U. Paquet and J. Tomczak and C. Zhang},
 pages = {104471--104520},
 publisher = {Curran Associates, Inc.},
 title = {Stepwise Alignment for Constrained Language Model Policy Optimization},
 url = {https://proceedings.neurips.cc/paper_files/paper/2024/file/bcfcf7232cb74e1ef82d751880ff835b-Paper-Conference.pdf},
 volume = {37},
 year = {2024}
}

@inproceedings{NEURIPS2023_91f18a12,
 author = {Zheng, Lianmin and Chiang, Wei-Lin and Sheng, Ying and Zhuang, Siyuan and Wu, Zhanghao and Zhuang, Yonghao and Lin, Zi and Li, Zhuohan and Li, Dacheng and Xing, Eric and Zhang, Hao and Gonzalez, Joseph E and Stoica, Ion},
 booktitle = {Advances in Neural Information Processing Systems},
 editor = {A. Oh and T. Naumann and A. Globerson and K. Saenko and M. Hardt and S. Levine},
 pages = {46595--46623},
 publisher = {Curran Associates, Inc.},
 title = {Judging LLM-as-a-Judge with MT-Bench and Chatbot Arena},
 url = {https://proceedings.neurips.cc/paper_files/paper/2023/file/91f18a1287b398d378ef22505bf41832-Paper-Datasets_and_Benchmarks.pdf},
 volume = {36},
 year = {2023}
}

@inproceedings{10.1145/3701716.3715265,
author = {Hu, Renjun and Cheng, Yi and Meng, Libin and Xia, Jiaxin and Zong, Yi and Shi, Xing and Lin, Wei},
title = {Training an LLM-as-a-Judge Model: Pipeline, Insights, and Practical Lessons},
year = {2025},
isbn = {9798400713316},
publisher = {Association for Computing Machinery},
address = {New York, NY, USA},
url = {https://doi.org/10.1145/3701716.3715265},
doi = {10.1145/3701716.3715265},
booktitle = {Companion Proceedings of the ACM on Web Conference 2025},
pages = {228–237},
numpages = {10},
location = {Sydney NSW, Australia},
series = {WWW '25}
}


\end{document}